\documentclass[conference]{IEEEtran}
\usepackage[utf8]{inputenc}
\usepackage[english]{babel}
\usepackage[hyphens]{url}
\usepackage[hidelinks]{hyperref}
\hypersetup{breaklinks=true}
\usepackage[numbers]{natbib}
\usepackage{amsthm}
\usepackage{amsmath} 
\usepackage{amssymb}

\theoremstyle{definition}
\newtheorem{definition}{Definition}[section]
\usepackage{soul}
\usepackage{multirow}
\usepackage{booktabs}
\usepackage{tabularx}
\usepackage{graphicx}
\usepackage{ tipa }
\usepackage{ dsfont }
\usepackage[dvipsnames]{xcolor}
\usepackage{tikz}
\usepackage{pgf}
\usepackage{pgfplots}
\usepackage{cancel}
\usepackage[inline]{enumitem}
\usepackage{stackengine}
\usepackage{rotating}
\usepackage{scalerel}
\usepackage[left=3cm,right=2cm,top=3cm,bottom=2cm]{geometry}
\usepackage{fancyhdr} 
\usepackage[colorinlistoftodos,prependcaption,textsize=tiny]{todonotes}
\usetikzlibrary{tikzmark,decorations.pathreplacing,calligraphy,arrows,shadows,petri,automata,shapes,shadows,trees,fit}
\tikzset{
	basic/.style  = {draw, text width=2cm, drop shadow, font=\sffamily, rectangle},
	root/.style   = {basic, rounded corners=2pt, thin, align=center,
		fill=green!30},
	level 2/.style = {basic, rounded corners=6pt, thin,align=center, fill=green!60,
		text width=8em},
	level 3/.style = {basic, thin, align=left, fill=pink!60, text width=8.5em}
}
\tikzset{myarrow/.style={-{Latex[length=2mm]}}}
\usepgfplotslibrary{groupplots}
\usetikzlibrary{shapes,arrows}
\usetikzlibrary{decorations.pathreplacing}
\usetikzlibrary{babel}
\usetikzlibrary{calc,arrows.meta,patterns,backgrounds}
\usetikzlibrary{positioning}
\usetikzlibrary{fit}

\tikzstyle{block}  =  [draw,  rectangle,  minimum  height=2em,  minimum  width=4em]
\tikzstyle{sum} = [draw, fill=blue!20, circle, node distance=1cm]
\tikzstyle{input} = [coordinate] \tikzstyle{output} = [coordinate]
\tikzstyle{pinstyle} = [pin edge={to-,thin,black}]
\usepackage[customcolors,norndcorners]{hf-tikz}
\pgfplotsset{compat=1.8}

\newcommand\Tstrut{\rule{0pt}{2.6ex}}         % = `top' strut
\newcommand\Bstrut{\rule[-0.9ex]{0pt}{0pt}}   % = `bottom' strut

\begin{document}

%\large %% FONT SIZE
\title{Learning Control Barrier Functions and their application in Reinforcement Learning: A Survey }

%\author{
%\IEEEauthorblockN{Maeva Guerrier}
%\IEEEauthorblockA{maeva.guerrier@polymtl.ca}
%\IEEEauthorblockN{Hassan Fouad}
%\IEEEauthorblockA{hassan.fouad@polymtl.ca}
%\IEEEauthorblockN{Hassan Fouad}
%\IEEEauthorblockA{hassan.fouad@polymtl.ca}
%}

\author{\IEEEauthorblockN{Maeva Guerrier}
	\IEEEauthorblockA{\textit{MISTLab} \\
		\textit{Polytechnique Montr\'eal}\\
		Montr\'eal, Canada \\
		maeva.guerrier@polymtl.ca}
	\and
	\IEEEauthorblockN{Hassan Fouad}
	\IEEEauthorblockA{\textit{MISTLab} \\
		\textit{Polytechnique Montr\'eal}\\
		Montr\'eal, Canada \\
		hassan.fouad@polymtl.ca}
	\and
	\IEEEauthorblockN{Giovanni Beltrame}
	\IEEEauthorblockA{\textit{MISTLab} \\
		\textit{Polytechnique Montr\'eal}\\
		Montr\'eal, Canada \\
		giovanni.beltrame@polymtl.ca}
	\and
%	\IEEEauthorblockN{4\textsuperscript{th} Given Name Surname}
%	\IEEEauthorblockA{\textit{dept. name of organization (of Aff.)} \\
%		\textit{name of organization (of Aff.)}\\
%		City, Country \\
%		email address or ORCID}
%	\and
%	\IEEEauthorblockN{5\textsuperscript{th} Given Name Surname}
%	\IEEEauthorblockA{\textit{dept. name of organization (of Aff.)} \\
%		\textit{name of organization (of Aff.)}\\
%		City, Country \\
%		email address or ORCID}
%	\and
%	\IEEEauthorblockN{6\textsuperscript{th} Given Name Surname}
%	\IEEEauthorblockA{\textit{dept. name of organization (of Aff.)} \\
%		\textit{name of organization (of Aff.)}\\
%		City, Country \\
%		email address or ORCID}
}

%\author{
%	\IEEEauthorblockN{Hassan Fouad}
%	\IEEEauthorblockA{hassan.fouad@polymtl.ca}
%}

\maketitle

\begin{abstract}
  Reinforcement learning is a powerful technique for developing new robot
  behaviors. However, typical lack of safety guarantees constitutes a hurdle for its practical application on real robots. 
%  As a result, reinforcement learning is mostly   confined to simulation, making it challenging to transfer learned policies to   real-world robots. 
  To address this issue, safe reinforcement learning aims to
  incorporate safety considerations, enabling faster transfer to real robots and
  facilitating lifelong learning.
  One promising approach within safe reinforcement learning is the use of
  control barrier functions. These functions provide a framework to ensure that
  the system remains in a safe state during the learning process. However,
  synthesizing control barrier functions is not straightforward and often
  requires ample domain knowledge. This challenge
  motivates the exploration of data-driven methods for automatically defining
  control barrier functions, which is highly appealing.
  We conduct a comprehensive review of the existing literature on
  safe reinforcement learning using control barrier functions. Additionally,
  we investigate various techniques for automatically learning the Control
  Barrier Functions, aiming to enhance the safety and efficacy of Reinforcement
  Learning in practical robot applications.
\end{abstract}

% keywords

% \fi
\IEEEpeerreviewmaketitle

\begin{table}
	\label{toa}
	\caption{Table of acronyms}
	\begin{center}
		% \begin{tabular}{|c|c|} 
			\begin{tabular}{p{0.12\textwidth}p{0.28\textwidth}}
				
%				\hline
				\textbf{Abbreviation} & \textbf{Meaning}\\
				\hline\Tstrut\Bstrut
				ML & Machine Learning\\\Tstrut\Bstrut
				RL & Reinforcement Learning\\\Tstrut\Bstrut
				SRL & Safe Reinforcement Learning\\\Tstrut\Bstrut
				CRL & Constrained Reinforcement Learning\\\Tstrut\Bstrut
				CMDP & Constrained Markov decision processes\\\Tstrut\Bstrut
				CBF & Control Barrier Function\\ \Tstrut\Bstrut
				RCBF & Robust Control Barrier Function\\\Tstrut\Bstrut
				ZBF & Zeroing Barrier Function\\\Tstrut\Bstrut
				SVM & Support Vector Machine\\\Tstrut\Bstrut
				PER & Prioritized Experience Replay\\\Tstrut\Bstrut
				HCBF & Hand-Crafted CBF\\\Tstrut\Bstrut
				ECBF & Exponential Control Barrier Function\\ \Tstrut\Bstrut
				QP & Quadratic Program\\\Tstrut\Bstrut
				HYCBF & Hybrid Control Barrier Function\\ \Tstrut\Bstrut
				LTL & Linear Temporal Logic\\\Tstrut\Bstrut
                LFD & Learning from demonstration\\\Tstrut\Bstrut
				DOBs & Disturbance Observers\\\Tstrut\Bstrut
				HJ & Hamitlon-Jacobi\\\Tstrut\Bstrut
				PACT & Perception-Action Causal Transformer\\\Tstrut\Bstrut
				LiDAR & Light Detection and Ranging\\\Tstrut\Bstrut
				CBVF & Control Barrier Value Function\\\Tstrut\Bstrut
				GP & Gaussian Process\\\Tstrut\Bstrut
				DP & Dynamic Programming\\\Tstrut\Bstrut
				CLBF & Control Lyapunov barrier Function\\\Tstrut\Bstrut
                GCBF & Generalized Control Barrier Function\\\Tstrut\Bstrut 
                UTCBF & Uncertainty-Tolerant Control Barrier Functions\\\Tstrut\Bstrut
                GCBF & Graph Control Barrier Function\\\Tstrut\Bstrut  
                GNN & Graph neural networks\\\Tstrut\Bstrut
				PACT& Perception-Action Causal Transformer\\
				\hline
			\end{tabular}
		\end{center}
	\end{table}

\section{INTRODUCTION} 

\subsection{Background}
In modern day robotics, one of the central quests that has captured the attention and interest of numerous researchers is to design  autonomous robotic systems that can operate in complex and highly dynamic environments. However, there are several challenges that impede the deployment of such robotic systems in real-life scenarios such as high degree of uncertainty in environments where robots operate~\cite{fan2020learning}, parameter errors and imperfections of models describing robots~\cite{abraham2020model}, and the increasingly complicated nature of some of these robotic systems~\cite{bhagat2019deep}. 

Although there is a sizable body of literature on model based methods that can deal with some aspects of these challenges such as adaptive~\cite{abiko2009adaptive,hung2008adaptive} and robust control~\cite{kolhe2013robust} methods, these methods remain limited in their ability of generalizing to more complicated environments and robot dynamics due to the relative difficulty of designing accurate models representative of robots and their environments.

Consequently, a lot of interest started shifting towards Machine Learning (ML) methods for designing control policies for robots due to their ability to abstract robot models and to handle system uncertainties through proper use of robot's data, which deem them effective methods for robot decision making \cite{KHOSRAVI2019311_ML_decision_making, Sahoo2020_ML_decision_making}. 

One specific flavor of ML that has been gaining much traction in the recent years for the purpose of designing policies for robots in presence of uncertainty is Reinforcement Learning (RL)~\cite{Sutton_RL_Intro}. The main premise of RL is to have an agent that learns behaviors through trial-and-error interactions with an uncertain environment~\cite{Sutton_RL_Intro}, such that the agent is incentivized to pick actions that maximize a certain reward through its interaction with the environment. This quest for reward maximization, i.e., exploitation, is coupled with a phase where the agent takes random actions to explore different parts of the solution space, i.e., exploration, with the aim of possibly finding other better policies on the long run, instead of greedily maximizing the reward and possibly getting stuck in local minima. 

This interplay between exploration and exploitation implies possible sacrifice of short term rewards, in favor of better long term rewards, and this is often called exploration/exploitation dilemma~\cite{Sutton_RL_Intro,efroni2020explorationexploitation}. Since the training of a RL agent takes place for a system in an uncertain environment, the policy coming out of the training process has an inherent capacity to handle uncertainties, but due to the exploration/exploitation dilemma it is possible that an agent takes unsafe actions while exploring its solution space, hoping for reward increase later. 

This potential for taking risky actions can be a serious limitation that limits the applicability of a learned policy in real-life~\cite{Challenges_Real_World_RL}, e.g. scenarios where safety is critical like robots interacting with humans~\cite{Provably_Safe_Deep_Reinforcement_Learning_for_Robotic_Manipulation_in_Human_Environments,Safe_Reinforcement_Learning_for_Single_Train_Trajectory_Optimization_via_Shield_SARSA}. Therefore, this necessitates the need for endowing the RL process with safety features to avoid safety violations during training and deployment. This can be generally challenging because deciding appropriate safety criteria may not be always straight forward, in addition to the fact that extracting safety margins, i.e., how far the system is from being unsafe, in systems with high-dimensional observations can introduce a lot of overhead.

Another motivation for the need to endow RL with safety is the fact that policy training in RL often takes place in simulated environments, which introduces an additional challenge due to potential mismatch between simulation and the real environment. This is often called simulation to reality gap~\cite{Sim_to_Lab_to_Real_Safe_Reinforcement_Learning_with_Shielding_and_Generalization_Guarantees}. This gap makes it more tricky to deploy policies trained in simulators to real-life environments, and mandates the presence of safety measures (during training and deployment) that are able to handle this gap.
%during deployment in real environments. 

This gives rise to the field of Safe Reinforcement Learning (SRL) that aims to augment RL processes with safety, typically through introduction of appropriate constraints that render RL policies safe during the training phase.
\subsection{Safety categorization in RL\label{subsec:category}}
According to~\cite{jacopo_paper}, there are three main types of constraints: \begin{enumerate*}
%	\item  Hard constraints ensure constraint satisfaction consistently. In this survey, we categorize papers as hard constraints when the employed method use filters and action monitoring during deployment of the agent. 
%	\item Soft constraints encourage the safe operation of the system. A typical way to achieve this is by using cost related constraints where constraints violations have a higher cost. This may come at the cost of unforeseen corner cases where the cost is inadequately defined, leading to unsafe scenarios. In this survey, we categorize papers that achieve safety without any form of monitoring over the agent's actions as methods employing soft constraints.  
%	\item Probabilistic constraints enforce constraints satisfactions up to a given probability. 	
	\item soft constraints that do not provide explicit safety guarantees, but rather encourage system states to behave in a safe manner 
	
%	\item hard constraints that use barriers to lead the learning process in a manner that enforces safety, 
	\item hard constraints which are actively enforced during the training phase, and ensure that system states reside in favorable portions of the state space
%	and rely on punishing the agent for proposing unsafe actions with large negative rewards or by constricting the action space.
	\item and probabilistic constraints that give a certain probability limit on violating safety, and can be seen as a middle ground between the previous two.	

\end{enumerate*}
% Figure~\ref{fig:safety_lvl} shows pictorial examples of hard, soft and probabilistic constraints. 

% \begin{figure}[hbt]
%     \centering
%     \includegraphics[keepaspectratio, width=90mm]{Figures/safety_level.pdf}
%     \caption{Visual representation of soft, probabilistic and hard constraints. Soft constraints do not provide explicit safety guarantees. Probabilistic constraints limit safety violation up to a given bound. Hard constraints directly enforce safety with the use of filter that monitor the system input. }
%     \label{fig:safety_lvl}
% \end{figure}

One common feature among soft constraint RL methods is that they do not explicitly alter an agent's action during exploration to prevent its state from being unsafe, but rather they aim to produce a policy that eventually satisfies some constraints. Examples of methods to achieve this include manipulating rewards~\cite{Reward_Constrained_Policy_Optimization}, approximate solutions of constrained MDP policy optimization with upper bounds on constraint violations~\cite{achiam2017constrained}, finding mixed strategy distributions to constrain a reward vector so it eventually converges to a desired set~\cite{Reinforcement_Learning_Convex_Constraints} and using weighted sums over multiple rewards~\cite{mannor2004geometric}.

%Typical RL approaches combine safety and optimization in a single policy by using large negative rewards when safety conditions are violated during training. However, this approach does not explicitly guarantee enforcing safety, thus giving rise to soft constraints \cite{Reward_Constrained_Policy_Optimization, Reinforcement_Learning_Convex_Constraints}. 
On the other hand, hard constraint RL methods aim to actively manipulate an agent's action during exploration to guarantee preventing its state from becoming unsafe. One popular way to achieve this manipulation is via introducing a safety "filter" or "shields" that permit an agent's action only if it is not leading to safety violations. Some examples of such methods include using CBFs~\cite{Safe_Reinforcement_Learning_for_an_Energy_Efficient_Driver_Assistance_System,Safe_Model_Free_Reinforcement_Learning_using_Disturbance_Observer_Based_Control_Barrier_Functions} to ensure forward invariance of safe sets, and safety shields that can estimate reachable sets to judge agent's potential of violating safety~\cite{Safe_Reinforcement_Learning_for_Control_Systems_A_Hybrid_Systems_Perspective_and_Case_Study,Safe_Reinforcement_Learning_via_Shielding,Safe_Multi_Agent_Reinforcement_Learning_via_Shielding}. 

%On the other hand, we can find several methods in the literature for introducing hard constraints, with the common theme of actively manipulating the policy action when it is expected to drive the system out of a certain safe set, thus effectively "filtering" the policy. Some examples include using two-player games in~\cite{Safe_Reinforcement_Learning_via_Shielding,Safe_Multi_Agent_Reinforcement_Learning_via_Shielding}, the use of Control Barrier Functions (CBFs)~\cite{Safe_Reinforcement_Learning_for_an_Energy_Efficient_Driver_Assistance_System}, and the approximate estimation of reachable sets in~\cite{Safe_Reinforcement_Learning_for_Control_Systems_A_Hybrid_Systems_Perspective_and_Case_Study}. 

Probabilistic guarantees can be introduced in several ways, such as using Gaussian Processes (GP) to model the system statistically
and then use a "safety filter"  setup to enforce constraints on that stochastic model with some probability. One example is in \cite{End_to_End_Safe_Reinforcement_Learning_through_Barrier_Functions_for_Safety_Critical_Continuous_Control_Tasks} that uses CBF constraints during the RL policy training, using a GP model. Backup policies are used in~\cite{work_proba_guarantee_Li_Shuo_Bastani} with a GP model trained from agent's interaction with environment, so they recover the state to a safe set around an equilibrium point in case the expected state violates safety.

%Probabilistic guarantees can be introduced in several ways, such as using Gaussian Processes (GP) to model the system statistically and apply CBF constraints on such model, while taking into account RL policy output during training~\cite{End_to_End_Safe_Reinforcement_Learning_through_Barrier_Functions_for_Safety_Critical_Continuous_Control_Tasks}. Another example of introducing probabilistic guarantees is presented  in~\cite{work_proba_guarantee_Li_Shuo_Bastani}, which proposes a RL framework for nonlinear systems with added stochastic noise, where additional backup policies are designed for recovering states to an invariant safe set around an equilibrium point, then activating these policies during RL training if the expected system state is unsafe under RL policy.   

%One typical challenge in RL and SRL that is present in many works previously mentioned, is the fact that policy training often takes place in simulated environments, which introduces an additional challenge due to potential mismatch between simulation, an approximation of reality itself, and the real environment, and is called simulation to reality gap~\cite{Sim_to_Lab_to_Real_Safe_Reinforcement_Learning_with_Shielding_and_Generalization_Guarantees}. This gap makes it more tricky to deploy policies trained in simulator to real-life environments.
So far we can notice that one common theme in safety methods in RL, through using safety filters and safety shields that aim to provide hard or probabilistic constraints, is the notion that safety is predicated on the ability to adjust RL policy to prevent the system from wandering to unsafe states, either deterministically or probabilistically. Our primary focus in this paper is surveying SRL methods and applications where CBF is the primary tool for ensuring safety (through providing guarantees on safe set invariance), both in training and deployment.

\subsection{CBF methods in RL}
CBF methods have been gaining significant amount of attention in the recent years as methods for certifying and guaranteeing safety in nonlinear systems~\cite{Control_Barrier_Functions_Theory_and_Applications}. This owes to their relative computational tractability, mission agnosticism and ability to provide guarantees on safety and forward invariance of safe sets, i.e. once a system starts in a safe set, it stays in this safe set. These attractive qualities lend CBFs as adequate tools to be utilized for safety in RL~\cite{Model_based_Reinforcemen_Learning_with_Provable_Safety_Guarantees_via_Control_Barrier_Functions, Model_based_Constrained_Reinforcement_Learning_using_Generalized_Control_Barrier_Function, End_to_End_Safe_Reinforcement_Learning_through_Barrier_Functions_for_Safety_Critical_Continuous_Control_Tasks,Reinforcement_Learning_Safe_Robot_Control_using_Control_Lyapunov_Barrier_Functions}.

With that in mind, CBF methods have some limitations that need some consideration to lend them more suitable for SRL. One such limitation is the fact that basic versions of CBFs are model-based, which need prior knowledge of the system model, both agent and environment, which affect their guarantees in scenarios where there is model uncertainty.\cite{Robust_Control_Barrier_Functions_Uncertainty_Estimation, Safe_Reinforcement_Learning_Using_Robust_Control_Barrier_Functions, Safe_Model_Free_Reinforcement_Learning_using_Disturbance_Observer_Based_Control_Barrier_Functions} Another limitation is the relative difficulty of crafting CBFs that encode desired safety for some applications, and the lack of \emph{a-priori} knowledge of the biggest available safe set associated with the desired safety specification \cite{Control_Barrier_Functions_Theory_and_Applications, Learning_Better_Control_Barrier_Function}.  

Although different methods exist for enabling CBFs to handle model uncertainty like Robust CBF~\cite{Robust_Control_Barrier_Functions_Uncertainty_Estimation, Robust_Control_Barrier_Functions_Nonlinear_Control_Systems_Uncertainty}, and adaptive CBF~\cite{lopez2020robust,taylor2020adaptive}, the latter issue motivates the use of learning methods to learn and formulate adequate CBFs based on system data~\cite{Learning_Barrier_Certificates_Towards_Safe_Reinforcement_Learning_with_Zero_Training_time_Violations,Learning_Barrier_Functions_for_Constrained_Motion_Planning_with_Dynamical_Systems,Learning_Control_Barrier_Functions_from_Expert_Demonstrations,Using_Reinforcement_Learning_to_Create_Control_Barrier_Functions_for_Explicit_Risk_Mitigation_in_Adversarial_Environments}, with the intention of lending CBFs more suitable for integration in RL context.

In this survey we highlight various methods for endowing RL with safety guarantees, while giving special attention to CBF based methods for safety in RL. Moreover, we shed some light on the body of literature related to learning CBFs from data, and the interplay between such methods with safety for RL during training and deployment.

\begin{figure}[hbt]
    \centering
    \includegraphics[keepaspectratio, width=84mm]{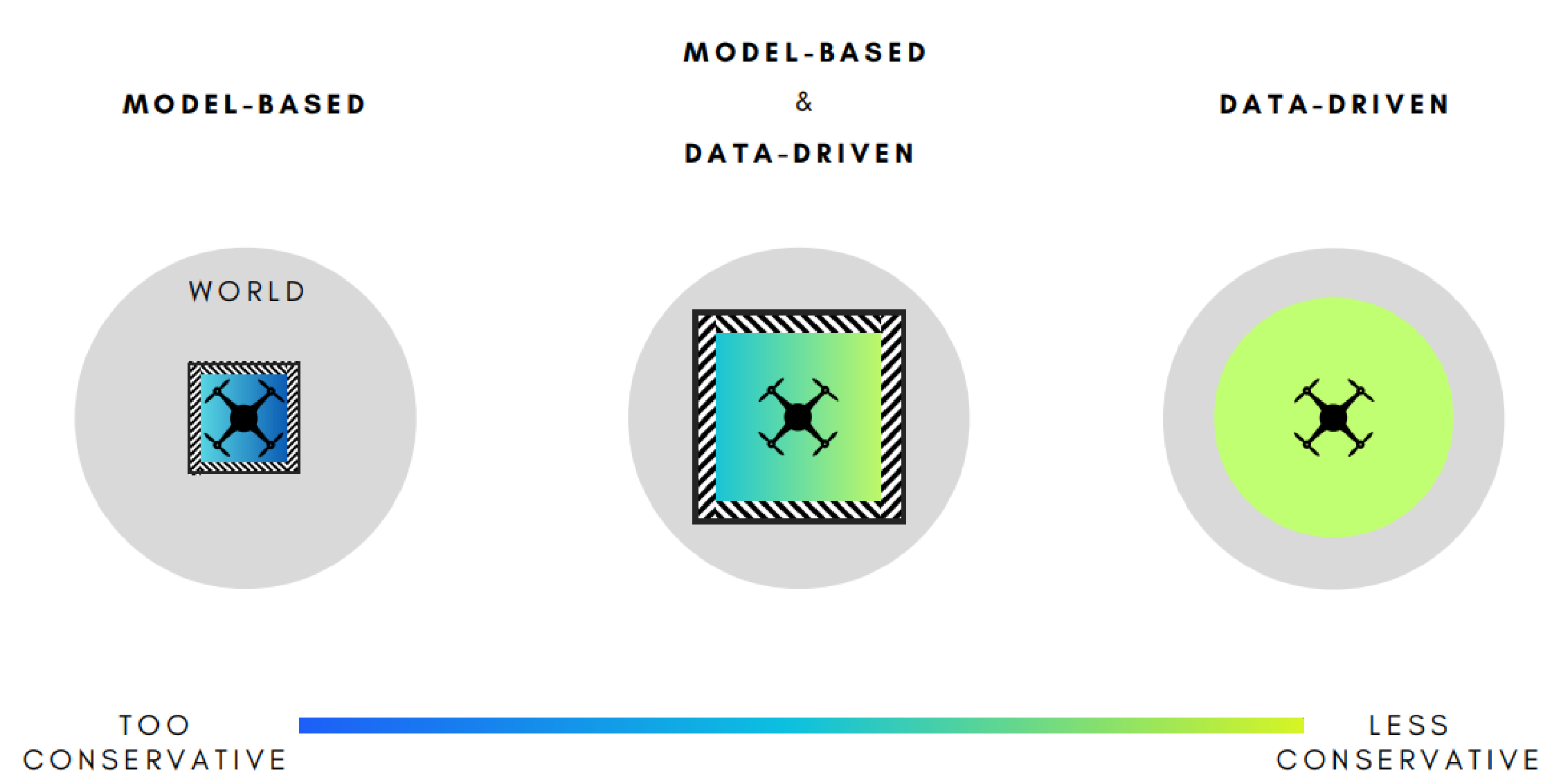}
    \caption{Illustrative depiction of model-based, data-driven and the combinaison of both approaches. Model-based approaches are robust, however, they are rigid and not flexible to unseen scenarios. Whereas, data-driven scenarios are adaptive to unforeseen cases. However, they are not bounded against unwanted cases. The combination of both methods allow to both have the flexibility of data-driven method while being able to enforce properties such as safety criteria. }
    \label{fig:mix_data_driven_model}
\end{figure}

\subsection{CONTRIBUTIONS}
The contributions of this survey are as follows:

\begin{itemize}
	\item We provide a comprehensive review of existing literature on safe reinforcement learning using CBF.

	\item We highlight the body of literature related to methods of learning CBFs from data
%    \item Discuss the relation of such method with SRL and how it can affect the policy generation.
%	\item We shed light on various methods of safety certification for RL. 

\end{itemize}

\subsection{ORGANIZATION} 

The organization of this paper is as follows. In section~\ref{sec:prelim} we introduce necessary preliminaries related to CBFs, RL and SRL, along with necessary definitions and concepts. In section~\ref{sec:SRL}, we cover safe reinforcement learning approaches in term of soft and hard constraints. Then we analyze different methods on how to construct CBF in section~\ref{sec:cbfconstruct} with a focus on learning from expert demonstrations and CBF enhancement methods. Section~\ref{sec:limits} presents a discussion concerning limitations of the methods and ideas we cover in the review, and we end with concluding remarks in section~\ref{sec:conc}.
%\newline

%\begin{center}
%\textbf{Table of abbreviations }
%\end{center}

\section{PRELIMINARIES\label{sec:prelim}}

\subsection{Control Barrier Functions} 
CBFs have become increasingly popular in the recent years as a tool for guaranteeing safety of dynamical systems~\cite{Control_Barrier_Functions_Theory_and_Applications}. In a CBF context, the definition of safety is associated with a continuously differentiable function $h(x)$, where a system is considered to be safe if $h(x) \geq 0$. Here $x\in\mathcal{D}\subset\mathbb{R}^n$ is the state of the system, and $\mathcal{D}\subset\mathbb{R}^n$ is a subset of $\mathbb{R}^n$ where the states reside . This function $h(x)$ can then be used to define the safe set $\mathcal{C}$ as~\cite{Control_Barrier_Functions_Theory_and_Applications}
\begin{equation}
	\begin{split}
		\mathcal{C} &= \{x\in\mathcal{D}: h(x) \geq 0\}\\
		Int(\mathcal{C})&=\{x\in\mathcal{D}: h(x) > 0\}\\
		\partial\mathcal{C}&=\{x\in\mathcal{D}: h(x) = 0\}
	\end{split}
	\label{eqn:safe_set}
\end{equation}
where $Int(\mathcal{C})$ is the interior of the safe set, and $\partial\mathcal{C}$ is its boundary. Therefore, maintaining safety implies the need for making the safe set $\mathcal{C}$ forward invariant, i.e., given that the system state start inside the safe set, they stay inside the safe set indefinitely. Based on the definition of $\mathcal{C}$ in \eqref{eqn:safe_set}, this implies the need to maintain $h(x)\geq 0$. 

Most of CBF methods are concerned with control affine systems, for reasons to be clarified shortly, that are defined as 
\begin{equation}
	\dot{x} = f(x) + g(x)u
\end{equation}
where $u\in U\subset\mathbb{R}^m$ is the system input, $U$ is the set of all admissible input values, and both $f(x)$ and $g(x)$ are two locally Lipschitz functions indicating system and input dynamics respectively.

%where $u\in U\subset\mathbb{R}^m$ is the system input, $U$ is the set of all admissible input values. Analogous to actions in RL, u signifies the choices or actions made within the system. Additionally, the terms $f(x)$ and $g(x)$ represent two locally Lipschitz functions. The function $f(x)$ characterizes the system dynamics, providing insights into how the system evolves over time. On the other hand, $g(x)$ describes the input dynamics, detailing how the system responds to various input values. 

One way to achieve the safety requirement $h(x)\geq 0$ is by manipulating the states such that $\dot{h}(x) \geq 0$, however this can be too restrictive, as it mandates the states to only change to conservatively increase the safety margin, which affects mission execution. Instead, a more popular way to maintain $h(x)\geq 0$ that is less restrictive is through achieving the following condition

%One way to achieve the safety requirement $h(x)\geq 0$ is by manipulating the states such that $\dot{h}(x) \geq 0$, however this can be too restrictive, as it mandates the states to only change to conservatively increase the safety margin, which affects mission execution. In the context of RL, this notion translates to imposing constraints on actions and necessitates a foreknowledge of the transition function, which is generally unavailable.A more popular way of maintaining $h(x)\geq 0$ that is less restrictive is through achieving the following condition
\begin{equation}
	\underbrace{L_fh + L_ghu}_{\dot{h}(x)}\geq -\alpha(h(x))
	\label{eqn:cbf_ineq}
\end{equation}
where $L_fh=\tfrac{\partial h}{\partial x}f$ and $L_gh=\tfrac{\partial h}{\partial x}g$ are the Lie derivatives of $h(x)$ in the direction of $f$ and $g$ respectively.

Several types of CBFs exist in the literature that provide safety guarantees for dynamical systems in different scenarios like having sampled systems~\cite{singletary2020control}, uncertainty in system parameters~\cite{lopez2020robust}, and uncertainty in system input~\cite{buch2021robust} to give a few examples. 
One of the popular version of CBFs that showcase the basic concepts of safe set invariance using CBFs is the Zeroing Control Barrier Functions (ZCBF), defined as follows

\begin{definition}
	For a subset $\mathcal{W}\subset\mathcal{C}\subset\mathcal{D}$, a continuously differentiable function $h(x)$ is a ZCBF if there exists an extended class $\mathcal{K}$ function $\alpha(h(x))$ such that 
	\begin{equation}
		\sup_{u \in U}\{L_fh(x)+L_gh(x)u\}\geq -\alpha(h(x)),\quad \forall x\in\mathcal{D}
		\label{eqn:cbf_cond}
	\end{equation} 
\end{definition}
Here $\alpha(h)$ belongs to extended class $\mathcal{K}$ functions if $\alpha(h)$ is strictly increasing and $\alpha(0)=0$. 
%\textit{Note: In control theory the class $\mathcal{K}$ function serves as a special comparison function to check the system stability.\newline} %confusing
We can also define a set of safe control inputs 
\begin{equation}
	K_{cbf}=\{u\in U: L_fh(x)+L_gh(x)u\geq -\alpha(h(x))\}
\end{equation}
which means that if $h(x)$ is a ZCBF, then any locally lipschitz function $u\in K_{cbf}$ will render the safe set $\mathcal{C}$ forward invariant, i.e., if $x(0)\in \mathcal{C} \Rightarrow x(t)\in\mathcal{C},\quad \forall t>0$.

Safe control input $u\in K_{cbf}$ discussed earlier is an input that can maintain $x\in\mathcal{C}$, but it is not necessarily the input that a system needs to carry out a desired task. Suppose a control action $k(x)\in U$ is a nominal control action for a system to carry out a task, one popular way to endow it with safety is through using a Quadratic Program (QP), where the CBF inequality \eqref{eqn:cbf_ineq} is a linear constraint in the QP

\begin{equation}
	\begin{split}
		u^*= &\underset{u\in U}{\arg\min} \text{ } \frac{1}{2}\parallel u - k(x) \parallel^2
		\\
		\underset{}{s.t.}& \text{ } L_{f}h(x) + L_{g}h(x)u \geq - \alpha h(x)
	\end{split}
	\label{eqn:qp_}
\end{equation}
Note the fact that the constraint in \eqref{eqn:qp_} is linear in $u$ due to the fact that the system is control affine in input, which makes the optimization problem a QP. This is advantageous owing to the existence of several numerically efficient methods for solving QP in real time. 

\subsection{Reinforcement learning}

\begin{figure}[hbt]
    \centering
    \includegraphics[keepaspectratio, width=84mm]{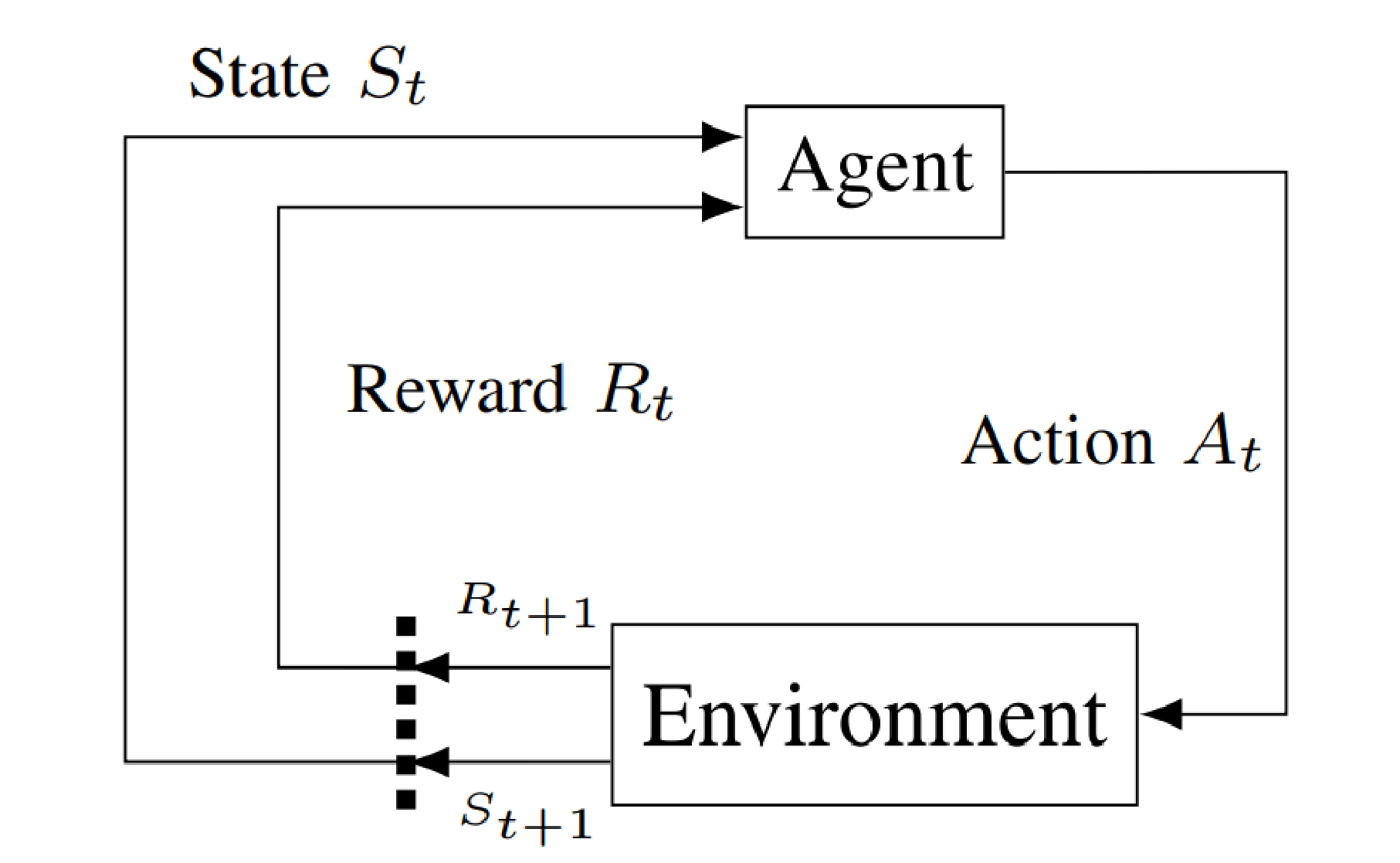}
    \caption{In reinforcement learning, an agent interacts with the environment. In a given state, the agent carries out an action and learns from the world by receiving a reward that is dependent on the previous state and the proposed action. The agent goes from one state to another with a state-transition probability that is unknown in real case scenarios.}
    \label{fig:RL_Agent_Interact_Env}
\end{figure}

RL is a goal-oriented learning method that aims to maximize rewards \cite{Sutton_RL_Intro}. 
An RL agent learns directly from interacting with the environment through observing the state $S_t$, applying an action $A_t$, then receiving a reward $R_{t+1}$ along with the new state $S_{t+1}$. This idea is depicted in Figure~\ref{fig:RL_Agent_Interact_Env}. Rewards can be positive, negative or neutral, i.e., zero. The agent then needs to adjust its action in an appropriate manner based on the received reward. 

In order to generate reasonable actions that achieve a desired task, an agent needs some knowledge about the relationship between actions it takes and consequent rewards it gets from the environment. Generally speaking, this relationship could be available in two ways~\cite{nguyen2019review}: \begin{enumerate*}
	\item through a model that describes the environment and helps the agent predict the relation between actions and rewards (model-based methods), 
	\item or the agent infers this relationship through previous values of action-reward pairs through repeated trials and interactions with the environment (model-free methods).
\end{enumerate*}

Model-based RL methods typically use the ability to predict future states and rewards to find adequate actions effectively. The presence of a model (readily available or learned from data) makes it applicable to find proper actions through planning~\cite{luo2022survey}, or the ability to augment already available data (resulting from interactions with the environment) with data generated from the model and then using model-free techniques~\cite{kurutach2018model}. This ability to predict the environment leads to enhancing sample efficiency for model-based methods, but on the other hand it can cause model biasing (i.e., good control performance with model, but poor results when used in real systems or slightly different models)~\cite{nguyen2019review}.

Model-free RL methods, on the other hand, rely primarily on interacting with the environment, through applying actions and receiving rewards to infer a relation between them, and use this knowledge directly to shape the actions needed to achieve a desired task. Such approach can accommodate more general systems without any prior knowledge of its model, thus solving the model biasing problem, but the tradeoff is that it needs more data samples (i.e., more interactions with the environment) to converge to a useful policy.

%One core feature of model-free RL methods is that the agent starts training with a limited knowledge of its environment and the effect of different actions on said environment. Therefore the state $S_t$ and reward $R_t$ signals play a crucial role in outlining this relationship, which in turn affects the choice of action later. To this end, there are generally two choices an agent can make:
In the model-free RL paradigm, an agent generally has two choices while interacting with the environment: 
\begin{itemize}
	\item either choose actions that help the agent better understand the underlying relationship between the environment and its action, i.e., prioritize exploration. This can be at the expense of sacrificing reward maximization to a certain degree, hoping to find a path toward better rewards in the long run. 
	\item Or, choose actions that maximize the agent's reward, given its current understanding of the system, i.e., prioritize exploitation. This can be at the expense of getting to local maxima, meaning it can get stuck with locally optimal actions, while potentially neglecting better actions not known to the agent owing to its poor exploration.
\end{itemize}

This is commonly called \emph{exploration-exploitation dilemma}~\cite{efroni2020explorationexploitation}, which mandates some savvy ways of striking a balance between how much an agent explores compared to how much it exploits. This implies the possibility of taking sub-optimal or unsafe actions during exploration, leading to undesirable behavior.
When deploying an agent into a robotic system safety concerns arise due to this dilemma. In a non-stationary environment, there is a need to learn continuously and thus explore. Hence the need of researching ways to guarantee safety.
%RL learns directly from environment's interactions (\textit{fig}.\ref{fig:RL_Agent_Interact_Env}) with the help of a stimulus \(R_{t}\): rewards either positive, negative or neutral (0). To gain knowledge of the environment, the agent has to choose to either exploit its current knowledge or explore to gain knowledge. 

%\newline 

The policy $\pi(s)$ is a mapping of states of the environment to actions that the agent can take for a given state. Typically in RL the agent does not have full knowledge of the environment and must take decisions under uncertainty. One popular tool to model such environments is the  \textit{Markov decision process} (MDP) described with the tuple $M = (S, S_{0}, A, P,R, \gamma)$, where $S$ is a finite set of states, $S_0$ is an initial state, $A$ is a finite set of actions, $P: S \times A \mapsto S$ is a probabilistic transition function, $R : S \times A \mapsto \mathbb{R}$ is a reward function, and $\gamma\in[0,1]$ is a discount factor.  

The cumulative reward $G_t$ is defined as
\begin{equation}
	G_t = \sum_{k=0}^{\infty}\gamma^kR_{t+k}
\end{equation}
and here we note that if $\gamma=0$, only immediate rewards are taken into account, while $\gamma=1$ means equally valuing all rewards, future and present.
%\begin{itemize}
%  \item A finite set of states \(S\), an initial state \(S_{0} \in S\).
%  \item A finite set of actions \(A\).
%  \item A probabilistic transition function \(P : S \times A \mapsto S \).
%  \item An immediate reward function \(R : S \times A \mapsto \mathbb{R}\)
%  \item The discount factor\footnote{The discount factor \(\gamma\), is a scalar value between 0 and 1, which represents the agent's preference for immediate rewards versus future rewards. A discount factor of 0 means the agent focuses on immediate rewards, while a discount factor of 1 means the agent equally values all future rewards.} \(\gamma\). In the case of a non-episodic MDP, where the time steps are infinite, the discount factor prevents an infinite sum of the cumulative rewards \(G_{t}\).
%%\newline
%\end{itemize}

An underlying key concept of RL is the Markov property. The Markov property states that the future is only dependent of the present, in other words, only the current state affects the next state:

\begin{equation}
P(S_{t+1} | S_{t}, A_{t}, S_{t-1}, A_{t-1}, ...)=P(S_{t+1} | S_{t}, A_{t})
\end{equation}
%\newline

The majority of RL algorithms operate under this assumption. The presence of the Markov property within an environment grants the agent the capacity to predict both the subsequent state and the anticipated reward, using only the information from the current state. 
The agent's goal is to find an optimal \textit{policy} $\pi^*$, which achieves the maximum cumulative reward $G_t$ from all sates~\cite{Sutton_RL_Intro}.

\begin{equation}
\pi^* = \underset{\pi}{\arg\max} \mathds{E}[G|\pi]
\end{equation}

%\newline
The value of a state $s$ under a policy $\pi$ is given by the state value function \( V_{\pi}(s) \), which corresponds to the expected return $G_t$ we can get if the agent starts at state $s$ and  follows the policy $\pi$. The state-value function $V$ is defined as:

\begin{equation}
V_{\pi}(s) = \sum_{a \in A(s)} \pi(a|s) \sum_{\substack{s' \in S,\\ r \in \mathds{R}}} p(s', r | s, a)(r + \gamma V_{\pi}(s'))
\end{equation}

The optimal value function is written as:  

\begin{equation}
V_{*}(s) = \underset{a \in A(s')}{\max} \sum_{\substack{s' \in S,\\ r \in \mathds{R}}} p(s', r | s, a)(r + \gamma V_{*}(s'))
\end{equation}

The value of a state $s$ given an action $a$ is given by the action value function, denoted as \( Q_{\pi}(s,a) \), which estimates the expected return $G_t$ the agent can obtain starting from state $s$, taking action $a$ and following the policy $\pi$.

\begin{equation}
Q_{\pi}(s,a) = \sum_{\substack{s' \in S,\\ r \in \mathds{R}}} p(s', r | s, a)[r + \gamma \sum_{a' \in A(s')} \pi(a'|s') Q_{\pi}(s',a')]
\end{equation}

The optimal action value function is denoted as \( Q_{*}(s,a) \) such that:

\begin{equation}
Q_{*}(s,a) = \sum_{\substack{s' \in S,\\ r \in \mathds{R}}} p(s', r | s, a)[r + \gamma \underset{a' \in A(s')}{\max} Q_{*}(s',a')]
\end{equation}

%\subsubsection{Model-based and model-free RL}
%There are two paradigms in RL; model-free and model-based RL \cite{Sutton_RL_Intro}. A model-based approach is any means that allow the agent to predict how the environment will react to its actions. Given an action-state pair the model produces the prediction of the next state and reward given this pair. If a model is deterministic it will only give a next-state-action pair given an action and a state. In the case where the model is stochastic, it will produces prediction over several next-states and their associated reward, with their state-transition probability. Those models are called distribution models \cite{Sutton_RL_Intro}. Model-based methods allow the generation of simulated experience and act as the environment itself. Model-base methods are sample efficient and thus can result in faster convergence, however a poor model of the environment will greatly hinder learning. Model-free methods simply consists of not modeling the environment and learning via trial and error by directly interacting with the environment. Model-free methods are less complex and are not prone to model bias in contrary to model-based approaches.

As stated earlier, one of the core ideas behind RL is the exploration-exploitation trade-off, which implies the possibility of exploring unfavorable actions that can make the agent's state unsafe.  
Typically in conventional RL methods, rewards do not explicitly incorporate safety, which limits the applicability of trained RL policies on physical system. 
This gives rise to SRL that focuses on learning optimal policies that satisfy safety  requirements during the learning and/or deployment phases.

\section{SAFE REINFORCEMENT LEARNING\label{sec:SRL}}
%\qquad
\subsection{SOFT CONSTRAINTS METHODS\label{sec:soft_const}} 
In this section, we give a brief overview of the literature related to SRL methods that adopt soft constraint, as defined in section~\ref{subsec:category}. We deliberately include such concise account for literature in this domain for completeness, but more detailed treatment of the literature can be found in~\cite{garcia2015comprehensive} and \cite{jacopo_paper}.

One common characteristic in such methods is that there are no explicit guarantees on safety during training, but rather they encourage policies to achieve safety in the long run. For example, \cite{kahn2017uncertainty} makes the RL policy searching risk aware by augmenting the cost function with a probabilistic risk model, and permits low impact safety violations to learn this model so it prevents high impact violations. 

Safety critics are presented  in~\cite{srinivasan2020learning}, where a pre-training phase aims to let a SAC agent explore unsafe states so it can train a safety critic and a policy, followed by a fine tuning phase where a new policy is trained while using the safety critic. 

Several methods depend on using CMDPs as a way of introducing constraints~\cite{achiam2017constrained,Reward_Constrained_Policy_Optimization, Reinforcement_Learning_Convex_Constraints} during policy construction.  In~\cite{achiam2017constrained} the authors present a method in which the cost function for policy optimization and the constraints are replaced with surrogate function based on upper bounds of objectives' divergence under policy change (both for main and constraint costs), which ensures monotonic improvement of policy with respect to constraints. In~\cite{Reward_Constrained_Policy_Optimization} the CMDP based policy optimization is tackled using Lagrange relaxation to produce a penalized reward function that is then used to train a "constrained" actor-critic, which converges to a policy that satisfies constraints.

%note that CPO uses trust region solver to find the policy while RCPO uses a lagrange relaxation

In previous works safety is encoded somehow in costs or rewards, but a different approach is taken 
in~\cite{Reinforcement_Learning_Convex_Constraints} where safety is defined as eventual convergence of a measurement vector to a desired set. To this end, the authors use a game theoretic framework to produce a mixed strategy probability distribution over a set of policies, which eventually decreases the distance between long term reward vector and the safe set.

Another approach to SRL is learning through demonstration. In~\cite{ConBaT_Control_Barrier_Transformer_for_Safety_Critical_Policy_Learning} a method is presented that uses expert demonstrations to  learn a safety critic that mimics the behavior of a CBF and augments a PACT architecture, such that the safety critic optimizes a proposed action to achieve safety. Although this method facilitates the labeling of demonstrations and has the ability to learn different safety definitions implicitly, it does not give guarantees on safety, and it might be vulnerable to out of distribution data or getting stuck at local minima in the action optimization phase.

One interesting approach is proposed in~\cite{Sim_to_Lab_to_Real_Safe_Reinforcement_Learning_with_Shielding_and_Generalization_Guarantees} which aims to reduce the effect of reality gap while adding safety through jointly training a backup policy, along with a performance policy and a discriminator, all conditioned on a latent variable to enhance generalization. The discriminator chooses the appropriate policy to avoid safety violations. 
%Both policies are constructed with SAC and are conditioned on a latent variable sampled from a distribution to encode diverse behaviors and enhance the ability to generalize. 
The policies obtained in the simulation phase are then fine tuned in a controlled environment setup to enhance  safety before deployment in real world. %if there is a limitation I would sy it would be the need for a lot of lab experiments, as they themselves mention in the paper

%the two policies are constructed with latent space states which allows for better distribution and generalization of policies.
%
%Another approach is the use of a backup policy as a filter \cite{Sim_to_Lab_to_Real_Safe_Reinforcement_Learning_with_Shielding_and_Generalization_Guarantees}. The backup policy intervene when the optimization policy fails to provide safe actions. In \cite{Sim_to_Lab_to_Real_Safe_Reinforcement_Learning_with_Shielding_and_Generalization_Guarantees} a shielding discriminator assess the safety of the actions proposed by the performance policy. In the case where the action is deemed unsafe the backup policy override the performance policy. The drawback is the need of learning both the shield as well as the backup policy. 
%train the learning agent \cite{ConBaT_Control_Barrier_Transformer_for_Safety_Critical_Policy_Learning}. However, providing exhaustive scenarios can be difficult, especially in complex tasks or environments.

\subsection{HARD and PROBABILISTIC CONSTRAINTS METHODS}

Unlike methods adopting soft constraints, methods enforcing hard safety constraints aim to give guarantees that ensure system safety throughout training and deployment, while probabilistic constraints qualify such guarantees to be within a certain desired probability range~\cite{jacopo_paper}. We note that such methods depend largely on having some form of model that describes the system, either known~\cite{Safe_Reinforcement_Learning_for_an_Energy_Efficient_Driver_Assistance_System} or learned~\cite{Reinforcement_Learning_Safe_Robot_Control_using_Control_Lyapunov_Barrier_Functions,Safe_Model_Free_Reinforcement_Learning_using_Disturbance_Observer_Based_Control_Barrier_Functions}.

\subsubsection{Using safety shields}
\begin{figure}[hbt]
    \centering
    \includegraphics[keepaspectratio, width=84mm]{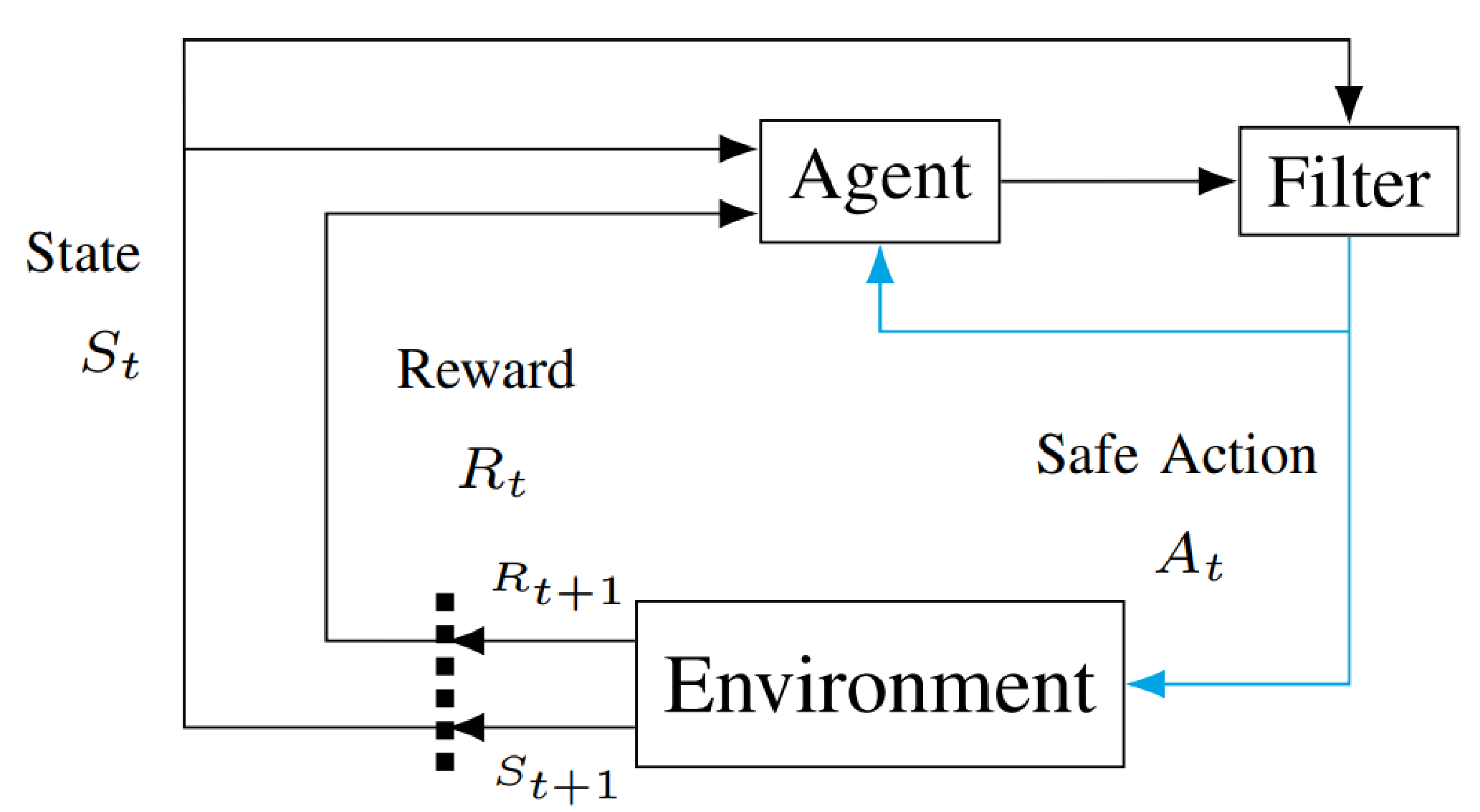}
    \caption{An illustrative example of how a filter function. The agent interact with the environment by proposing an action at a given state and shift to the next state given an unknown state-transition probability. The action is monitored by the filter while training and on deployment. The filter maps unsafe actions to safe ones to maintain system safety. If the filter has to change the action, the agent is aware of that fact and learns form its mistake, otherwise, it directly receives the reward from the environment.}
    \label{fig:filter_fig}
\end{figure}

One idea that has garnered much attention by many researchers recently is the use of safety shields~\cite{Safe_Reinforcement_Learning_via_Shielding,Safe_Multi_Agent_Reinforcement_Learning_via_Shielding,Safe_Reinforcement_Learning_for_Control_Systems_A_Hybrid_Systems_Perspective_and_Case_Study, Sim_to_Lab_to_Real_Safe_Reinforcement_Learning_with_Shielding_and_Generalization_Guarantees, Safe_Reinforcement_Learning_for_Single_Train_Trajectory_Optimization_via_Shield_SARSA} Figure~\ref{fig:filter_fig} shows a schematic of a basic safety shield architecture. Herein, we use the notion of a safety filter and a safety shield interchangeably. The core idea of as safety  shield is that it monitors an action before being applied to the system, and it modifies this action only in case it is expected to violate safety.
%Safety filters provide additional safety as they assess the safety of the proposed action in real-time and aim to prevent the agent from falling in unsafe cases. Policies that are subjects to filtering concepts can be seen as RL approaches with hard constraints.

The work in \cite{Safe_Reinforcement_Learning_via_Shielding} introduces an unsafe action filtration based on the concept of a finite-sate reactive system or \textit{Mealy machine} \cite{mealy_machine} and \textit{Linear Temporal Logic}. The main idea is to construct a two player game between the system and environment, and use it to estimate a winning region, i.e., subset of state space where states can always be recovered to safety by a backup controller. The actions are then minimally modified and punishment signals are used to enhance the RL policy training.  The main limitation in \cite{Safe_Reinforcement_Learning_via_Shielding} is that it only considers discrete sets and spaces, whereas real-life scenarios are in a continuous space.

The same shield construct is extended to a multi-agent setting by \cite{Safe_Multi_Agent_Reinforcement_Learning_via_Shielding}, which presents a centralized shield that monitors the states of all agents. This implies that the problem grows in complexity with more robots. To remedy this issue, the authors also introduce a set of factorized shields, each monitoring a subset of the state space such that each agent is covered by at least one factorized shield. Agents switch shields near boundary states between these shields to ensure coordination.

%that monitors a subset of agents. The centralized shield coordinates and checks the output of the factored shields in term of compatibility. However, the problem only has been shifted as computational complexity will increase with the number of factored shields. %unclear from the paper

The method in \cite{Safe_Reinforcement_Learning_for_Control_Systems_A_Hybrid_Systems_Perspective_and_Case_Study} adopts a similar framework as \cite{Safe_Reinforcement_Learning_via_Shielding}, but it decreases the complexity by learning a function as the shield directly, which predicts a binary outcome (safe/unsafe) based on observations. In order  to reach the goal, a planner system combined with the shield proposes intermediary goals that are inherently safe.

Another example of using shields is presented in \cite{Safe_Reinforcement_Learning_for_Single_Train_Trajectory_Optimization_via_Shield_SARSA}, where a shield is combined with the SARSA algorithm \cite{Sutton_RL_Intro}. Upon choosing an action that is deemed unsafe by the shield, the action is removed from the action set and the exploration strategy is used to pick another action.
In a similar fashion, the authors of \cite{work_proba_guarantee_Li_Shuo_Bastani} have embedded the shield concept within a policy, that overrides the current action upon constraint violation. The shield policy infers safety by checking the next state degree of recoverability. 

%This idea of inherently enforcing safety is also done in \cite{Safe_Reinforcement_Learning_for_an_Energy_Efficient_Driver_Assistance_System} by leveraging a filtering approach using an Exponential Control Barrier Function (ECBF). 

%Another approach is the use of a backup policy as a filter \cite{Sim_to_Lab_to_Real_Safe_Reinforcement_Learning_with_Shielding_and_Generalization_Guarantees}. The backup policy intervene when the optimization policy fails to provide safe actions. In \cite{Sim_to_Lab_to_Real_Safe_Reinforcement_Learning_with_Shielding_and_Generalization_Guarantees} a shielding discriminator assess the safety of the actions proposed by the performance policy. In the case where the action is deemed unsafe the backup policy override the performance policy. The drawback is the need of learning both the shield as well as the backup policy. 

\subsubsection{USING CBF SAFETY FILTERS\label{sec:cbf_rl}}
%\qquad
%\newline

%There are two outlined main approaches for safe reinforcement learning: learning safety during training only or leveraging filtering methods to enforce safety during deployment as well.
%\newline
Control Barrier Functions have gained a lot of attention in the past decade and have become very popular tools for enforcing safety constraints on dynamical systems~\cite{Control_Barrier_Functions_Theory_and_Applications}. Over the past few years we notice a consistent increase in the domains of applications of such tool, but in this paper we are interested in CBFs from an ML and RL perspective.

Most common CBF formulations are model-based in nature, implying the need for a model to estimate CBF gradients and be able to derive safe actions. However, in an RL context it is common to not have \emph{a-priori} known system models. Consequently, several CBF based SRL implementations use uncertainty-aware CBFs. Typically, such methods use a nominal system model, and treat differences from the actual system as disturbances.

% we can talk about Safe_Model_based_Reinforcement_Learning_with_Stability_Guarantees in the previous section
Several works \cite{End_to_End_Safe_Reinforcement_Learning_through_Barrier_Functions_for_Safety_Critical_Continuous_Control_Tasks,cheng2023safe}, to cite a few, use CBFs to enforce safety during the training of RL agents. 
The authors of \cite{End_to_End_Safe_Reinforcement_Learning_through_Barrier_Functions_for_Safety_Critical_Continuous_Control_Tasks} proposes a method for augmenting a generic RL model-free method with safety during training using CBFs safety filters. Their method depend on a nominal system model with a GP to model the disturbance, and such framework allows for having probabilistic guarantees. Moreover, they propose a method to incentivize the trained policy to produce safer actions through incorporating the history of previously filtered actions. 
\textcolor{black}{In line with the approach of using GP to learn disturbance, the authors of \cite{Safe_Reinforcement_Learning_Using_Robust_Control_Barrier_Functions} utilize the learned disturbance estimates to make subtle adjustments to the actions of their policy, ensuring system safety with minimal alterations. } 

%In the other-hand \cite{Safe_Model_based_Reinforcement_Learning_with_Stability_Guarantees}  presented an RL algorithm to ensure high-probability safety guarantees for policies taking inspiration from asymptotic stability in control theory instead.

Disturbance observers (DOBs) provide effective tools to deal with uncertainties~\cite{cheng2023safe}. In this work, the authors present a high relative degree CBF formulation with disturbance, and adopt a few assumptions about the boundedness of the disturbance and its derivative. This enables formulating a predictor for the upper bound of disturbance and incorporate such function in the CBF constraint that filters RL actions during training. The safe action and consequent reward are then added to a buffer that is used in the RL training process to enhance the final policy. 

\textcolor{black}{Addressing uncertainties, this paper \cite{Model_based_Reinforcemen_Learning_with_Provable_Safety_Guarantees_via_Control_Barrier_Functions} introduces Uncertainty-Tolerant Control Barrier Functions (UTCBFs), a novel CBF category that accommodates model uncertainty while ensuring provable safety guarantees. For systems featuring unknown dynamics, model dynamics and state predictions are derived using GP. Initially, the UTCBF constraint is conservative during early exploration due to a large uniform error bound caused by model uncertainty. As more data is gathered, the error bound converges, gradually relaxing the restrictiveness of the constraint. The policy is updated through constraint optimization.}

\textcolor{black}{One notable limitation identified by \cite{Model_based_Constrained_Reinforcement_Learning_using_Generalized_Control_Barrier_Function} is CBF inapplicability to dynamic systems with high relative degrees. To address this issue, the authors introduce the Generalized Control Barrier Function (GCBF), a modification designed specifically for handling constraints high degree systems. The GCBF restricts policy updates through a model-based constrained policy optimization approach, effectively mitigating the challenges posed by high relative-degree dynamics.
The authors of \cite{Safe_Reinforcement_Learning_for_an_Energy_Efficient_Driver_Assistance_System} also acknowledge that traditional CBFs may not be suitable for systems with high relative degree. In this context, customized CBFs, specifically exponential barrier functions (ECBFs), designed to handle high relative degree systems, are employed. The RL agent's proposed traction torque actions undergo a safety filtering process using ECBFs before being applied to the system.}

%The authors of \cite{Safe_Model_Free_Reinforcement_Learning_using_Disturbance_Observer_Based_Control_Barrier_Functions} took another approach by using disturbance observers (DOBs) incorporated with CBFs eliminating the concern of uncertain dynamics.

\section{CBF CONSTRUCTION\label{sec:cbfconstruct}}

From the discussion in section~\ref{sec:cbf_rl} we notice that generally speaking, methods utilizing CBFs to enforce safety for RL policies need to accommodate uncertainties~\cite{End_to_End_Safe_Reinforcement_Learning_through_Barrier_Functions_for_Safety_Critical_Continuous_Control_Tasks,Safe_Model_Free_Reinforcement_Learning_using_Disturbance_Observer_Based_Control_Barrier_Functions}. Moreover, in many scenarios constructing the CBF for RL policy training is challenging~\cite{ConBaT_Control_Barrier_Transformer_for_Safety_Critical_Policy_Learning,Safe_Inverse_Reinforcement_Learning_via_Control_Barrier_Function}. To that end, it is of order to give a brief review of different methods that aim to construct and adapt CBFs using data, which we find applicable and relevant to CBF application in RL domains.

\begin{table*}[htbp!]
\caption{\textbf{Overview of SRL methods focused on CBFs.} Online: real-time stream of data. Offline: Static dataset.}\label{tab_SRL}
\begin{tabular}{|p{0.17\textwidth}|c|c|p{0.09\textwidth}|p{0.09\textwidth}|p{0.09\textwidth}|p{0.2\textwidth}|}
\hline
\multicolumn{1}{|c|}{\textit{\textbf{References}}} & \multicolumn{1}{|c|}{\textit{\textbf{Deployed}}} & \multicolumn{1}{|c|}{\textit{\textbf{Model-base}}} & \multicolumn{1}{|c|}{\textit{\textbf{Model-free}}} & \multicolumn{1}{|c|}{\textit{\textbf{Online}}} & \multicolumn{1}{|c|}{\textit{\textbf{Offline}}}  & \multicolumn{1}{c|}{\textit{\textbf{Guarantees}}}                                                                     

\tabularnewline \hline \Tstrut
\cite{Safe_Reinforcement_Learning_for_Single_Train_Trajectory_Optimization_via_Shield_SARSA, Safe_Multi_Agent_Reinforcement_Learning_via_Shielding, Safe_Model_Free_Reinforcement_Learning_using_Disturbance_Observer_Based_Control_Barrier_Functions, Safe_Reinforcement_Learning_for_an_Energy_Efficient_Driver_Assistance_System, Provably_Safe_Deep_Reinforcement_Learning_for_Robotic_Manipulation_in_Human_Environments}& &  & \checkmark & \checkmark & & Hard constraints 
\tabularnewline \hline\Tstrut\Bstrut
\cite{Sim_to_Lab_to_Real_Safe_Reinforcement_Learning_with_Shielding_and_Generalization_Guarantees, work_proba_guarantee_Li_Shuo_Bastani}& \checkmark &  & \checkmark & \checkmark & & Hard constraints 
\tabularnewline \hline\Tstrut\Bstrut
\cite{Safe_Inverse_Reinforcement_Learning_via_Control_Barrier_Function} &  &  & \checkmark &  & \checkmark & Soft constraints 
\tabularnewline \hline\Tstrut\Bstrut
\cite{Reinforcement_Learning_Safe_Robot_Control_using_Control_Lyapunov_Barrier_Functions} &   &  & \checkmark & \checkmark &  & Soft constraints 
\tabularnewline \hline\Tstrut\Bstrut
\cite{Safe_Reinforcement_Learning_Using_Robust_Control_Barrier_Functions} &   & \checkmark & \checkmark & \checkmark &  & Soft constraints 
\tabularnewline \hline\Tstrut\Bstrut
\cite{Model_based_Reinforcemen_Learning_with_Provable_Safety_Guarantees_via_Control_Barrier_Functions} &   & \checkmark &  & \checkmark &  & Probabilistic constraints 
\tabularnewline \hline\Tstrut\Bstrut
\cite{End_to_End_Safe_Reinforcement_Learning_through_Barrier_Functions_for_Safety_Critical_Continuous_Control_Tasks} &   &  & \checkmark & \checkmark &  & Probabilistic constraints  
\tabularnewline \hline\Tstrut\Bstrut
\cite{Model_based_Constrained_Reinforcement_Learning_using_Generalized_Control_Barrier_Function} & \checkmark & \checkmark &  & \checkmark &  & Probabilistic constraints 

\tabularnewline \hline
\end{tabular}
\end{table*}

\subsection{CBF CONSTRUCTED WITH MACHINE LEARNING METHODS}
\subsubsection{RL based methods}
Different ML can be used to design and infer CBFs from data. One such popular direction is via using RL. For example, the method from~\cite{Safe_Inverse_Reinforcement_Learning_via_Control_Barrier_Function} uses inverse RL, which originally infers a reward function from data. An expert provides demonstration data for safe and unsafe actions, and a CBF is constructed as a neural network. Loss functions are designed to mimic the behavior of an actual CBF, and the minimization of which leads to a CBF that can discriminate between safe and unsafe sets.

In a fashion similar to~\cite{Safe_Inverse_Reinforcement_Learning_via_Control_Barrier_Function}, the framework from~\cite{chuchuPaper} constructs a CBF through proper use of adequate loss functions that enforce properties of a CBF. However, it differs from~\cite{Safe_Inverse_Reinforcement_Learning_via_Control_Barrier_Function} in that instead of using explicit expert demonstrations as input data, it collects data from system rollouts using the current policy from the recent episode, then updates the policy and recollect data,i.e., it adopts an on-policy training strategy. Moreover, it applies methods for rendering observations quantity and permutation invariant, and it adopts a setup inspired from a CBF-QP framework~\cite{Control_Barrier_Functions_Theory_and_Applications} to refine the policy.

The work in \cite{Reinforcement_Learning_Safe_Robot_Control_using_Control_Lyapunov_Barrier_Functions} aims to learn CLBFs, which are functions similar to CBFs, albeit with different properties that ensure both safety and tracking simultaneously. The method is based on modifying the Bellman equation of the state value function such that it respects properties of a CLBF. The core idea is to learn the state value function under CLBF constraints and find the policy according to that function. The learned policy is used as a critic in an actor-critic setting. 

In \cite{Using_Reinforcement_Learning_to_Create_Control_Barrier_Functions_for_Explicit_Risk_Mitigation_in_Adversarial_Environments} the authors use the RL value function to construct a CBF and directly incorporates safety in rewards. The rewards are only given at the end of an episode, however, this implies that the agent cannot learn from intermediate failures and successes.

The authors of \cite{Synthesis_Control_Barrier_Functions_Using_Supervised_Machine_Learning_Approach} created a framework that constructs a CBF based on LiDAR data. The information on the environment retrieved by the LiDAR outlines the set of safe and unsafe samples. 
The CBF is synthesize as a two hidden layer Gaussian Kernel neural network, the first being a kernel machine and the second a SVM.

The work in \cite{zhang2024gcbf+} presents a distributed framework for safe multi-agent scenarios. They introduce the Graph Control Barrier Function framework denoted GCBF+, utilizing graph neural networks (GNNs) to parameterize a candidate GCBF. To enable goal-reaching behavior, a goal node and edges between each agent and their respective goal are incorporated into the input features. Training involves minimizing the sum of the CBF loss for both the GCBF and the distributed controller. 
Alternative methodologies \cite{Sim_to_Lab_to_Real_Safe_Reinforcement_Learning_with_Shielding_and_Generalization_Guarantees, Safe_Multi_Agent_Reinforcement_Learning_via_Shielding, Safe_Reinforcement_Learning_for_Single_Train_Trajectory_Optimization_via_Shield_SARSA} often grapple with a conflicting reward-penalty framework, primarily because, in unsafe scenarios, the backup or nominal policy override the optimal one, leading to suboptimal behavior. However, in this case, their \cite{zhang2024gcbf+} loss formulation integrates both safety and optimization within a unified policy, effectively sidestepping the trade-off dilemma between collision avoidance and goal reaching.

Safety critics could be used to construct CBF-like functions from expert demonstrations to endow RL policies with safety on the long run \cite{ConBaT_Control_Barrier_Transformer_for_Safety_Critical_Policy_Learning}. Such methods have relatively simple structure and allows for easy labeling of demonstration data, however, they are vulnerable to getting stuck in local minima during the policy optimization phase as discussed in section~\ref{sec:soft_const}.
\subsubsection{Learning CBF from demonstrations}

Another set of popular techniques for CBF construction is through learning from demonstrations. The work in
\cite{Learning_Barrier_Functions_for_Constrained_Motion_Planning_with_Dynamical_Systems} introduces a method for estimating a set of ZCBFs, each modeled as a linear combination of states, from incoming training data. This framework adopts a method of incrementally clustering input points into subsets of unsafe states, then fits a number of ZCBFs for each of these clusters, and uses these fitted functions to enforce safety on the system. 

%integrated both learning from demonstrations and iterative learning of the CBF. The authors defined their ZCBF as a linear function where the main idea is to find a low-dimensional embedding which represents the linear ZCBF for each cluster points. 

In \cite{Learning_Control_Barrier_Functions_from_Expert_Demonstrations} the CBF is synthesized by learning from a dataset of safe and unsafe trajectories.  Given the set of expert trajectories, data points are extracted such that state-action pairs present safe behaviors from demonstrations and with those points a safe set is learned.

Unlike other methods that use both safe and unsafe demonstrations~\cite{Safe_Inverse_Reinforcement_Learning_via_Control_Barrier_Function,ConBaT_Control_Barrier_Transformer_for_Safety_Critical_Policy_Learning}, the work in \cite{Learning_Hybrid_Control_Barrier_Functions_from_Data} presents a way for constructing a CBF for hybrid systems while using only demonstrations of safe behaviors. Instead of having explicit unsafe demonstrations, the authors define an unsafe set outside of safe demonstrations and use constrained optimization to find a CBF, modeled as DNN, that satisfies CBF properties, i.e. equations~\eqref{eqn:safe_set} and~\eqref{eqn:cbf_cond}. 
%However, the formulated problem, although valid for safety criteria, leads to a more conservative CBF.
%\newline

The framework from ~\cite{Learning_Robust_Output_Control_Barrier_Functions_from_Safe_Expert_Demonstrations} extends the results from ~\cite{Learning_Hybrid_Control_Barrier_Functions_from_Data} by taking into account model and state estimation errors. To that end, a Robust Output Control Barrier Function (ROCBF) is constructed from expert demonstrations through constrained optimization in a similar fashion to ~\cite{Learning_Hybrid_Control_Barrier_Functions_from_Data}, with similar definitions of safe and unsafe regions. The ROCBF can be then used to enforce safety due to its ability to account for model and measurement uncertainties.

GPs are used in~\cite{khan2022gaussian} to model and construct CBFs from demonstration data. In this work, Gaussian CBFs are presented, which aims to produce a posterior with a positive value for regions in state space with high confidence. To that end, data samples from the safe sets directly shapes the CBF structure, which becomes negative in areas where less data is collected (due to being unsafe). Such structure allows for great flexibility in CBF formulation, as well as online construction and enforcement of safety with these CBFs.

\textcolor{black}{The authors of \cite{CBF_Unknown_Nonlinear_Sys_using_GP} undertake the learning process of a GP and its associated confidence while employing Satisfiability Modulo Theories solvers to find a valid control barrier functions and the corresponding control policies within the input space. This learned GP aid in comprehending the unknown control affine nonlinear dynamics and addressing uncertainties inherent in a learned model for the control barrier functions. The work in \cite{Gaussian_CBF} also addresses the limitations of CBF when confronted with model uncertainties. The proposed method in~\cite{Gaussian_CBF} tackles this challenge by acquiring insights into the unmodeled dynamics through the utilization of GP. It integrates the Gaussian CBF approach and formulates it within the context of an online QP optimization problem.
}

\textcolor{black}{	
Furthermore, the authors \cite{Safety_Uncertainty_CBF_using_GP} tackle as well the challenge of model uncertainties and their influence on CBF. These uncertainties are captured by calculating the posterior variance through GP, conditioned on previous measurement states. They integrate a Gaussian CBF, incorporating safety uncertainty by utilizing the posterior variance derived from past measurement data. The system input is compared against the Gaussian-CBF framed as a QP problem, and adjustments are made if necessary. }

\textcolor{black}{Expanding upon the same concern in a decentralized multi-robot context, \cite{Decentralized_Robust_Collision_Avoidance_Coop_Multirobot_GP_CBF} integrates the assessment of individual model uncertainty through GP models. This ensures that each robot can ensure safety with a high level of confidence. The formulation of safe controllers involves solving a QP problem, constrained by decentralized robust CBF conditions, guided by the GP uncertainty estimates.
Another strategy involves acquiring knowledge about the attraction region directly for nonlinear systems. For example, the methodology presented in \cite{learn_attract_reg_GP} relies on non-parametric Bayesian frameworks like GP and Bayesian Optimization to assess and broaden the safe set.}

%To do so, they solve a constrained optimization problem by using neural networks to estimate the CBF. In a parallel fashion to \cite{Learning_Better_Control_Barrier_Function}, safety criteria are labelled data points, safe and unsafe, based on demonstrations. They identify boundaries of the set and label the said bounds as unsafe, which result in a constrained optimization problem. 

\subsubsection{Using CBF priors}

One main feature of basic CBFs is the fact that they are myopic in nature, meaning that they only care about enforcing safety regardless of mission execution. Additionally, the extent of the safe set to which the system states are confined is implicitly implied by the choice of CBF candidates, which might be too conservative and not reflect a potentially bigger true safe set of states.

To that end, several methods exist in the literature that aim to enhance the system performance and feasibility~\cite{onlineCBF_decentralize_multi_agent_nav,Learning_Differentiable_Safety_Critical_Control_CBF_Generalization_Novel_Environments}, as well as finding CBF candidate with expanded safe sets~\cite{designCBF_from_state_constraints,Learning_Better_Control_Barrier_Function}. The main idea behind these methods is to start by some known CBF with some tunable factors, which are trained using different ML techniques to achieve the aforementioned goals.

The authors of \cite{yang2024learning} suggest a learning-enabled method for developing local CBFs to ensure the safety of diverse nonlinear hybrid dynamical systems. Their goal is to tackle the curse of dimensionality in high-dimensional systems. In their approach, they contemplate the incorporation of both safe and unsafe system configurations when updating the CBF. The CBF is directly adjusted based on the dynamically computed new unsafe set using dynamic programming and is employed to guide hybrid systems.

The work in~\cite{onlineCBF_decentralize_multi_agent_nav} addresses the issue of potential reduction in system performance and lack of feasible solutions (i.e. potentially conflicting constraints, or system getting stuck due to lack of safe actions that progress the mission) with a fixed extended class $\mathcal{K}$ function in the CBF constraint structure, i.e., a fixed $\alpha(h)$ function in \eqref{eqn:cbf_ineq}. The proposed framework represents the $\alpha$ function as a GNN and trains an RL policy that regulates $\alpha$ to control how conservative or aggressive the constraint behavior is. This policy aims to increase the chance of having feasible solutions, while enhancing the overall system performance and generalizability to new environments.

%The authors of \cite{onlineCBF_decentralize_multi_agent_nav} fine tune CBFs based on the agent observations using parameterized policies. The propose CBF is not fixed and adapt to non-stationary environment.
Similarly in~\cite{Learning_Differentiable_Safety_Critical_Control_CBF_Generalization_Novel_Environments}, the authors aim to enhance applicability of CBF constraints to new environments, as well as system performance. However, they use Exponential CBFs to account for systems with high relative degree. The framework laid out in~\cite{Learning_Differentiable_Safety_Critical_Control_CBF_Generalization_Novel_Environments} trains an agent that manipulates the CBF constraint structure at higher relative degree, in a manner similar to~\cite{onlineCBF_decentralize_multi_agent_nav}, to eventually enhance system performance, reduce control effort and enhance feasibility.

%The authors of \cite{Learning_Differentiable_Safety_Critical_Control_CBF_Generalization_Novel_Environments} leverages another approach by using an Exponential Control Barrier Function based Quadratic Program (ECBF-QP) to handle novel environments. The ECBF-QP is embedded in a neural network as an additional layer. This process enable the search of a looser CBF constraint flexible due to the data-driven nature of neural networks. 

Besides the different methods of constructing CBFs we discussed previously, an additional approach is through the refinement of ``less accurate'' CBFs. For instance,~\cite{Learning_Better_Control_Barrier_Function} starts by using a handcrafted CBF (HCBF), along with its safe set, as an initial guess for the true safe set. This candidate is then augmented with an increment that aims to bring the CBF closer to the true, possibly bigger, safe set. The increment is modeled with a Deep Differential Network (DDN) and trained based on sampled data for safe and unsafe states. 

This framework was further extended in~\cite{Data_Efficient_Control_Barrier_Function_Refinement}, which proposes a method for sampling data points based on Prioritized Experience Replay, leading to a better sample efficiency and needing less samples to train the CBF increment.

The main theme in~\cite{Learning_Better_Control_Barrier_Function} and~\cite{Data_Efficient_Control_Barrier_Function_Refinement} is that we start by a conservative estimation of a safe set that is smaller than its true volume and the goal is to expand it. Conversely, in~\cite{designCBF_from_state_constraints} the main theme is to start from a bigger set, which is not forward invariant and defined by a set of constraints, and shrink it to converge to the true safe set. To do so, an approximate CBF representation is obtained from the original constraints through scaling and offsetting the state variables, then safe and unsafe data points are then sampled to help modifying the scale and offset of states in a manner that ``trims'' the bigger set and makes it converge to the true safe and forward invariant set.

%An additional tactic is the refinement of a first hand-crafted CBF (HCBF) with machine learning methods.The paper \cite{designCBF_from_state_constraints} enhance a HCBF candidate by expanding the safe set in a data-driven fashion with the use of uniform and non-uniform scaling and offset.

In \cite{Learning_Barrier_Certificates_Towards_Safe_Reinforcement_Learning_with_Zero_Training_time_Violations}, the authors leverage an approach that iteratively learns barrier certificates with the help of a prior HCBF. However, the authors did not address the issue of having a limited amount of data, 
%whereas \cite{Data_Efficient_Control_Barrier_Function_Refinement} confronted the challenge of learning CBFs using scarce data. The underlying dilemma is the generation of a CBF that covers the "true" safe set. In order to have more coverage of the safe states while having insufficient, the authors leverage data Prioritized Experience Replay (PER). 

%The authors of \cite{Learning_Better_Control_Barrier_Function} also discussed the concern of having a less conservative approach and extending the coverage of safe states. Unlike \cite{Data_Efficient_Control_Barrier_Function_Refinement}, \cite{Learning_Better_Control_Barrier_Function} uses a combination of Model Predictive Control based approaches and CBFs to cover a larger portion of the "true" safe set. 

None of the above works acknowledge the issue of updating an invalid CBF. \cite{Refining_Control_Barrier_Functions_through_Hamilton_Jacobi_Reachability} addressed this concern with a framework that allows the refinement of invalid and overly conservative CBFs. The limit of their work is the need of knowing the obstacles dynamics. 
The table \ref{tab_cbf_learn} gives an overview of the CBF learning methods discussed above.

\begin{table*}[htbp!]
\caption{\textbf{Overview of CBF learning methods.} Online: real-time stream of data. Offline: Static dataset. }\label{tab_cbf_learn}
\begin{tabular}{|c|c|c|c|c|c|c|p{0.42\textwidth}|}
\hline
\multicolumn{1}{|c|}{\textit{\textbf{References}}} & \multicolumn{1}{|c|}{\textit{\textbf{Deployed}}} & \multicolumn{1}{|c|}{\textit{\textbf{Online}}} & \multicolumn{1}{|c|}{\textit{\textbf{Offline}}} & \multicolumn{1}{|c|}{\textit{\textbf{LFD}}} & \multicolumn{1}{|c|}{\textit{\textbf{ML}}}  & \multicolumn{1}{c|}{\textit{\textbf{CBF Prior}}} & \multicolumn{1}{c|}{\textit{\textbf{Comments}}}
\tabularnewline \hline \Tstrut
\cite{Learning_Robust_Output_Control_Barrier_Functions_from_Safe_Expert_Demonstrations, ConBaT_Control_Barrier_Transformer_for_Safety_Critical_Policy_Learning, Learning_Control_Barrier_Functions_from_Expert_Demonstrations} & & & \checkmark & \checkmark & & & These works rely on learning CBF through expert demonstrations.
\tabularnewline \hline \Tstrut\Bstrut
\cite{Learning_Barrier_Functions_for_Constrained_Motion_Planning_with_Dynamical_Systems} & \checkmark & \checkmark & & \checkmark & & & Incorporates a kinesthetic teaching strategy, the authors opted for a flexible addition or removal of learned constraints.
\tabularnewline \hline \Tstrut\Bstrut
\cite{Learning_Hybrid_Control_Barrier_Functions_from_Data} &  & \checkmark & & \checkmark & & & Tackles the expansion of safe sets in the context of CBF under the influence of model uncertainty.
\tabularnewline \hline \Tstrut\Bstrut
\cite{onlineCBF_decentralize_multi_agent_nav} & \checkmark & \checkmark & &  & \checkmark & & Expands upon the application of CBF, their study further incorporates their utilization to guarantee safety in a decentralized manner within a multi-robot system.
\tabularnewline \hline \Tstrut\Bstrut
\cite{Using_Reinforcement_Learning_to_Create_Control_Barrier_Functions_for_Explicit_Risk_Mitigation_in_Adversarial_Environments, Synthesis_Control_Barrier_Functions_Using_Supervised_Machine_Learning_Approach, Learning_Differentiable_Safety_Critical_Control_CBF_Generalization_Novel_Environments} & & \checkmark & & & \checkmark & & The authors express a concern regarding the acquisition of a broadly applicable CBF through machine learning methodologies.
\tabularnewline \hline \Tstrut\Bstrut
\cite{Safe_Reinforcement_Learning_for_an_Energy_Efficient_Driver_Assistance_System} & & \checkmark & & & \checkmark & & Considers safety as a priority in the context of reinforcement learning, moreover, the authors give due consideration to the preservation of energy.
\tabularnewline \hline \Tstrut\Bstrut
\cite{Safety_Uncertainty_CBF_using_GP} & \checkmark & \checkmark & & & \checkmark & & This study addresses the model's uncertainty by leveraging real-time measurements of system states, while incorporating the uncertainty measure into a Gaussian process.
\tabularnewline \hline \Tstrut\Bstrut
\cite{Decentralized_Robust_Collision_Avoidance_Coop_Multirobot_GP_CBF, Gaussian_CBF, CBF_Unknown_Nonlinear_Sys_using_GP, learn_attract_reg_GP} & & \checkmark & & & \checkmark & & Uses Gaussian Processes to learn the underlying model uncertainty and incorporate it in the CBF formulation.
% Addresses a pivotal concern in the application of CBF, the authors focus on handling model uncertainties, which are effectively addressed through Gaussian Process techniques to learn the underlying uncertainty in the model.
\tabularnewline \hline \Tstrut\Bstrut
\cite{AlwaysSafe_Reinforcement_Learning_without_Safety_Constraint_Violations_during_Training}  & & & \checkmark & & & \checkmark & The objective of  this study is to achieve a flawless training process with no safety violations by progressively enhancing a minimal CBF through iterative development and refinement.
\tabularnewline \hline \Tstrut\Bstrut
\cite{Learning_Barrier_Certificates_Towards_Safe_Reinforcement_Learning_with_Zero_Training_time_Violations, Refining_Control_Barrier_Functions_through_Hamilton_Jacobi_Reachability, Learning_Better_Control_Barrier_Function, Data_Efficient_Control_Barrier_Function_Refinement}  &  &\checkmark & & & &\checkmark & The authors strive to acquire an improved CBF through data based on a minimal CBF, all the while ensuring that safety constraints remain unbreached.
\tabularnewline \hline \Tstrut\Bstrut
\cite{Reinforcement_Learning_Safe_Robot_Control_using_Control_Lyapunov_Barrier_Functions} & \checkmark & \checkmark & & & & \checkmark & Employs a blend of control barrier and Lyapunov function strategies to build upon a minimal valid CBF.
\tabularnewline \hline
\end{tabular}
\end{table*}

\section{Discussion}
\subsection{MODEL-FREE VS MODEL-BASED}
%<<<<<<< HEAD
%From the discussion in section~\ref{sec:SRL} we notice that in general the ability to provide hard or probabilistic safety guarantees is contingent on having some form of system model, either known \emph{a-priori}~\cite{Safe_Reinforcement_Learning_for_an_Energy_Efficient_Driver_Assistance_System,End_to_End_Safe_Reinforcement_Learning_through_Barrier_Functions_for_Safety_Critical_Continuous_Control_Tasks, Safe_Reinforcement_Learning_for_Single_Train_Trajectory_Optimization_via_Shield_SARSA}, or learned~\cite{Model_based_Reinforcemen_Learning_with_Provable_Safety_Guarantees_via_Control_Barrier_Functions, End_to_End_Safe_Reinforcement_Learning_through_Barrier_Functions_for_Safety_Critical_Continuous_Control_Tasks, Safe_Reinforcement_Learning_Using_Robust_Control_Barrier_Functions}.
%%\todo{add more citations for both categories}. 
%Moreover, the strength of guarantees depend directly on the fidelity of the adopted model. Some methods ameliorate model imperfections, that can potentially cause safety violations, through making assumptions about worst case disturbances~\cite{Safe_Model_Free_Reinforcement_Learning_using_Disturbance_Observer_Based_Control_Barrier_Functions}, learn such disturbances during training~\cite{Safe_Reinforcement_Learning_Using_Robust_Control_Barrier_Functions}, or use sampling~\cite{work_proba_guarantee_Li_Shuo_Bastani}. 
%=======
From the discussion in section~\ref{sec:SRL} we notice that in general the ability to provide hard or probabilistic safety guarantees is contingent on having some form of system model, either known \emph{a-priori}~\cite{Safe_Reinforcement_Learning_for_an_Energy_Efficient_Driver_Assistance_System,End_to_End_Safe_Reinforcement_Learning_through_Barrier_Functions_for_Safety_Critical_Continuous_Control_Tasks, Safe_Reinforcement_Learning_for_Single_Train_Trajectory_Optimization_via_Shield_SARSA, Learning_Barrier_Functions_for_Constrained_Motion_Planning_with_Dynamical_Systems}, or learned~\cite{Model_based_Reinforcemen_Learning_with_Provable_Safety_Guarantees_via_Control_Barrier_Functions, End_to_End_Safe_Reinforcement_Learning_through_Barrier_Functions_for_Safety_Critical_Continuous_Control_Tasks, Safe_Reinforcement_Learning_Using_Robust_Control_Barrier_Functions, CBF_Unknown_Nonlinear_Sys_using_GP, Safety_Uncertainty_CBF_using_GP}. Moreover, the strength of guarantees depend directly on the fidelity of the adopted model. Some methods ameliorate model imperfections, that can potentially cause safety violations, through making assumptions about worst case disturbances~\cite{Safe_Model_Free_Reinforcement_Learning_using_Disturbance_Observer_Based_Control_Barrier_Functions}, learn such disturbances during training~\cite{Safe_Reinforcement_Learning_Using_Robust_Control_Barrier_Functions}, or use sampling~\cite{work_proba_guarantee_Li_Shuo_Bastani}.

We also notice that in general, model-based RL methods can be more convenient to produce hard and probabilistic guarantees, especially with CBFs~\cite{Model_based_Constrained_Reinforcement_Learning_using_Generalized_Control_Barrier_Function,Model_based_Reinforcemen_Learning_with_Provable_Safety_Guarantees_via_Control_Barrier_Functions,Safe_Reinforcement_Learning_Using_Robust_Control_Barrier_Functions} due to existence of system models that can help predicting future states for a given action, in addition to estimating gradients of CBFs, which allows formulating safe actions. However, the same observation holds true for systems using model-free RL methods but with an added model specifically for the  purpose of ensuring safety~\cite{Safe_Model_Free_Reinforcement_Learning_using_Disturbance_Observer_Based_Control_Barrier_Functions,Safe_Reinforcement_Learning_via_Shielding,Safe_Reinforcement_Learning_for_an_Energy_Efficient_Driver_Assistance_System,work_proba_guarantee_Li_Shuo_Bastani}.

Methods that use actor-critics  are highly dependable on the learned critic, if the latter does not properly take into account safety this will hinder the final policy. 
One of the shortcomings of both approaches \cite{Using_Reinforcement_Learning_to_Create_Control_Barrier_Functions_for_Explicit_Risk_Mitigation_in_Adversarial_Environments,Reinforcement_Learning_Safe_Robot_Control_using_Control_Lyapunov_Barrier_Functions} is the exploitation-exploration dilemma of RL, that still allow the agent to brake safety criteria within a small margin. \textit{A response to that problem could be to only act greedily however this could lead to local optimum.} 

Model-based RL methods have an advantage regarding sample efficiency in comparison to model-free RL methods, however, they have problems with model bias and poor generalisability~\cite{Sutton_RL_Intro}. Although model-free methods do not suffer model bias, they are mostly associated with soft constraints, provided they are not augmented with system models specifically for safety~\cite{kahn2017uncertainty,srinivasan2020learning,achiam2017constrained,Reward_Constrained_Policy_Optimization}. This owes to the lack of predictive power from models, and limiting knowledge of environment on immediate interactions.

Due to the complete independence from models in model-free RL, some of these methods might permit safety violations to learn safety on the long run~\cite{kahn2017uncertainty,srinivasan2020learning}, approximate solutions of CMDPs~\cite{achiam2017constrained} which might cause safety violations, or encoding the constraints in the reward~\cite{Reward_Constrained_Policy_Optimization}, aiming to converge to safe behaviors asymptotically.

\subsection{Effect of safety on policy performance}

Adding safety into account during training of RL policies has several effects on the overall outcome of these policies. We are interested in giving some insights into the effect of such safety consideration on three main aspects, namely \begin{enumerate*}
	\item sample efficiency and speed of convergence,
	\item policy overall safety
	\item and final task execution quality.
\end{enumerate*}
We are first shed some light on the effect of adding different types of safety constraints (soft, hard, probabilistic) on the aforementioned metrics, and contrast that with overall performance of RL methods that do not consider safety. Moreover, we  compare between the overall performance of safe RL methods enforcing different strengths of constraints in the light of the previously mentioned metrics.

One thing we can notice consistently is the increased sample efficiency and speed of policy convergence, as well as policy safety, of different safe RL methods applying hard~\cite{Safe_Reinforcement_Learning_Using_Robust_Control_Barrier_Functions,Safe_Reinforcement_Learning_via_Shielding}, probabilistic~\cite{End_to_End_Safe_Reinforcement_Learning_through_Barrier_Functions_for_Safety_Critical_Continuous_Control_Tasks,Model_based_Constrained_Reinforcement_Learning_using_Generalized_Control_Barrier_Function} and soft~\cite{ConBaT_Control_Barrier_Transformer_for_Safety_Critical_Policy_Learning} constraints in comparison to their non-safe counterparts. For example, the results in~\cite{Safe_Reinforcement_Learning_Using_Robust_Control_Barrier_Functions} show an increased sample efficiency of a SAC augmented with robust CBF during training. The results in~\cite{Safe_Reinforcement_Learning_via_Shielding} show an improvement both in convergence speed and safety during training for the proposed shielded RL agent compared to the unshielded one. Original TRPO~\cite{schulman2015trust} and DDPG~\cite{lillicrap2015continuous} policies were compared to their safety augmented counterparts in~\cite{End_to_End_Safe_Reinforcement_Learning_through_Barrier_Functions_for_Safety_Critical_Continuous_Control_Tasks} and the results show a significant improvement in both sample efficiency and safety. Similarly,~\cite{Model_based_Constrained_Reinforcement_Learning_using_Generalized_Control_Barrier_Function} shows better adherence to safety and convergence speed of the modified MBPO compared to an original MBPO~\cite{janner2019trust}.

Also introducing soft constraints to the RL learning process leads to an enhanced safety and sample efficiency as in~\cite{ConBaT_Control_Barrier_Transformer_for_Safety_Critical_Policy_Learning} that shows an added advantage in achieving safety compared to original version of PACT~\cite{bonatti2023pact}. The typical improvement in sample efficiency could be attributed to the fact that adding safety constraints, especially hard and probabilistic, trims down the available action space, and consequently the state space, leading the agent to explore less ``non-useful'' parts of the state space, while being guided towards achieving the goal.

In addition to enhancing sample efficiency and safety performance, adding safety during training also in many occasions has an advantage in terms of the overall reward during training~\cite{End_to_End_Safe_Reinforcement_Learning_through_Barrier_Functions_for_Safety_Critical_Continuous_Control_Tasks,ConBaT_Control_Barrier_Transformer_for_Safety_Critical_Policy_Learning,Safe_Reinforcement_Learning_via_Shielding}.

\subsection{Challenges and limitations\label{sec:limits}}

Herein, we provide a brief discussion of challenges in the field of safe RL in the light of the discussions presented so far, as well as potential opportunities for future research.
\subsubsection{Effect of sim2real gap on  safety}
Throughout this study we notice that most of the development of safe RL policies takes place mostly in simulated environments, much like normal RL. The inevitable mismatch between simulation models and real environments can be a source of several issues regarding the applicability and validity of safe RL policies.

Some popular solutions that we can find in the literature to deal with this issue include domain randomization~\cite{ranaweera2023evaluation} and optimization of simulator parameters~\cite{kaspar2020sim2real}. However, there are few works, e.g.~\cite{Safe_Reinforcement_Learning_for_an_Energy_Efficient_Driver_Assistance_System}, that deal with certification of safe RL policies during deployment in a fashion that takes into account this reality gap. The certification we find in literature is mostly during  training in the context of hard/probabilistic constraints. There are few attempts to tackle this issue (safety certification during deployment) in the context of soft constraints, e.g.~\cite{Sim_to_Lab_to_Real_Safe_Reinforcement_Learning_with_Shielding_and_Generalization_Guarantees}, but not for more strict constraints.

Regular safety filtration, using robust and adaptive CBF methods for example, can be used as an external safety layer applied to the output of the policy. However, such methods can be arbitrarily conservative in a manner that does not match the conservativeness during training, which can affect the overall performance. We think that this aspect of safety certification of RL methods during deployment could be a future direction of research.
\subsubsection{Generalizability and zero shot performance}
It is worth mentioning that one important aspect that can asses the overall quality of safe RL policies is their zero-shot performance, i.e. the validity of RL policies when deployed in different simulated environment from the one used during training. We notice that few works consider this aspect of assessment. Some of the few examples that give some insight to zero shot \cite{zero_shot_survey, zero_shot_task_GEN_DPRL} performance include~\cite{Safe_Reinforcement_Learning_Using_Robust_Control_Barrier_Functions}

\section{Conclusion\label{sec:conc}}
Reinforcement learning is a powerful tool that gained attention in the past couple of decades due to their significant versatility. Such potential will benefit from methods to endow RL policies with safety to enhance the training process and bolster their applicability in real life. In this short review we focus on safe RL methods that attempt to provide policies with safety features, with a specific interest in methods that use CBFs as a part of safety filters during training.

We outline several types of safety constraints in literature and show where CBF based methods fall, and we give some examples of works using CBFs for the purpose of safety in RL. We also give an outline of the literature related to learning CBFs from data, since this is a common aspect in the use of CBFs for safe RL. Lastly, we discuss some insights from our observation of the literature.

We believe that the field of safe RL, and more specifically the use of CBFs for safety certification of RL methods, will keep getting more attention in the years to come, and researchers will tackle several issues that include linking certification in training with deployment to reduce the effect of the sim2real gap, devise more sample efficient methods for building CBFs from data, and improve the generalization and zero shot performance of safe RL methods.

%Reinforcement learning is a powerful tool that addresses sequential decision making under uncertainties, which allows the development of new robot behaviors. However, its practical application to real robots is limited due to a lack
%of safety guarantees. Safe reinforcement learning addresses this issue by incorporating safety concerns. One promising approach is the use of control barrier functions to enforce hard constraints and enable faster transfer to real robots. On their own, control barrier functions can be too conservative, which hinder their flexibility against unforeseen scenarios. Furthermore, they are non-trivially handcrafted. This motivates the need of data driven approaches to learn control barrier functions. Researchers leverage different approaches to learn  control barrier functions, such as learning from expert demonstrations and enhancing control barrier functions using machine learning methods. The incorporation of safety in safe reinforcement learning and by extension in reinforcement learning, is done in two main ways. Learning under a safety construct that monitor and correct unsafe actions during the training process and the other way, uses it as a filter during deployment. The main outline of the current state of the literature, is the use of control barrier functions during the learning process to enforce safety guarantees. An interesting emerging approach is the integration of control barrier functions properties directly into the reinforcement learning problem. 

% Add space between references
\setlength{\bibsep}{7pt plus 0.3ex}
\bibliography{main}
\end{document}